\documentclass{llncs}
\usepackage{graphicx}
\usepackage{verbatim}
\usepackage{multirow}
\usepackage{booktabs}
\usepackage{xcolor}
\usepackage{amsmath}

\newcommand{\xx}{\vec{x}}
\newcommand{\beq}{\begin{equation}}
\newcommand{\eeq}{\end{equation}}
\newcommand{\mr}[1]{\mathrm{#1}}

\begin{document}

\title{\Large{Dense Multi-path U-Net for Ischemic Stroke Lesion  Segmentation in Multiple Image Modalities}}


\author{Jose Dolz\inst{1}, Ismail Ben Ayed\inst{1}, Christian Desrosiers\inst{1}}

\institute{Laboratory of Imaging, Vision and Artificial Intelligence\\
Ecole de techologie sup\'erieure, Montreal, Canada\\
\email{jose.dolz@livia.etsmtl.ca}}

\maketitle

Delineating infarcted tissue in ischemic stroke lesions is crucial to determine the extend of damage and optimal treatment for this life-threatening condition. However, this problem remains challenging due to high variability of ischemic strokes' location and shape. Recently, fully-convolutional neural networks (CNN), in particular those based on U-Net \cite{ronneberger2015u}, have led to improved performances for this task \cite{choi2016ensemble}. In this work, we propose a novel architecture that improves standard U-Net based methods in three important ways. First, instead of combining the available image modalities at the input, each of them is processed in a different path to better exploit their unique information. Moreover, the network is densely-connected (i.e., each layer is connected to all following layers), both within each path and across different paths, similar to HyperDenseNet \cite{dolz2018hyperdense}. This gives our model the freedom to learn the scale at which modalities should be processed and combined. Finally, inspired by the Inception architecture \cite{szegedy2016rethinking}, we improve standard U-Net modules by extending inception modules with two convolutional blocks with dilated convolutions of different scale. This helps handling the variability in lesion sizes. We split the 93 stroke datasets into training and validation sets containing 83 and 9 examples respectively. Our network was trained on a NVidia TITAN XP GPU with 16 GBs RAM, using ADAM as optimizer and a learning rate of 1$\times$10$^{-5}$ during 200 epochs. Training took around 5 hours and segmentation of a whole volume took between 0.2 and 2 seconds, as average. The performance on the test set obtained by our method is compared to several baselines, to demonstrate the effectiveness of our architecture, and to a state-of-art architecture that employs factorized dilated convolutions, i.e., ERFNet \cite{romera2018erfnet}.

\section{Introduction}

Stroke is one the leading causes of global death, with an estimate of 6 million cases each year \cite{lopez2006global,seshadri2007lifetime}. It is also a major cause of long-term disability, resulting in reduced motor control, sensory or emotional disturbances, difficulty understanding language, and memory deficit. Cerebral ischemia, which comes from the blockage of blood vessels in the brain, represents approximately 80\% of all stroke cases \cite{sudlow1997comparable,feigin2003stroke}. Brain imaging methods based on Computed Tomography (CT) and Magnetic Resonance Imaging (MRI) are typically employed to evaluate stroke patients \cite{van2007acute}. Early-stage ischemic strokes appear as a hypodense regions in CT, making them hard to locate with this modality. MRI sequences, such as T1 weighted, T2 weighted, fluid-attenuated  inversion  recovery  (FLAIR), and  diffusion-weighted imaging (DWI), provide a clearer image of brain tissues than CT, and are preferred modalities to assess the location and evolution of ischemic stroke lesions \cite{chalela2007magnetic,barber2005imaging,lansberg2000comparison}.

The precise delineation of stroke lesions is critical to determine the extend of tissue damage and its impact on cognitive function. However,  manual segmentation of lesions in multi-modal MRI data is time-consuming as well as prone to inter and intra-observer variability. Developing methods for the automatic segmentation can thus contribute to having more efficient and reliable tools to quantify stroke lesions over time \cite{praveen2018ischemic}. Over the years, various semi-automated and automated techniques have been proposed for segmenting lesions \cite{rekik2012medical,maier2015classifiers}. Recently, deep convolutional neural networks (CNNs) have shown high performance for this task, outperforming standard segmentation approaches on benchmark datasets \cite{maier2017isles,winzeck2018isles,guerrero2018white,chen2017fully,kamnitsas2017efficient}. 

Multi-modal image segmentation based on CNNs is typically addressed with an \emph{early fusion} strategy, where multiple modalities are merged from the original input space of low-level features \cite{zhang2015deep,moeskops2016automatic,kamnitsas2015multi,kamnitsas2017efficient,dolz2017deep,valverde2017improving}. This strategy assumes a simple relationship (e.g., linear) between different modalities, which may not correspond to reality \cite{Srivastava14}. For instance, the method in \cite{zhang2015deep} learns complementary information from T1, T2 and FA images, however the relationship between these images may be more complex  due to the different image acquisition processes. To better account for this complexity, Nie et al. \cite{nie2016fully} proposed a \emph{late fusion} approach, where each modality is processed by an independent CNN whose outputs were fused in deep layers. The authors showed this strategy to outperform early fusion on the task of infant brain segmentation.

More recently, Aig\"{u}n et al. explored different ways of combining multiple modalities \cite{aygun2018multi}. In this work, all modalities are considered as separate inputs to different CNNs, which are later fused at an `early', `middle' or `late' point. Although it was found that `late' fusion provides better performance, as in \cite{nie2016fully}, this method relies on a single-layer fusion to model the relation between all modalities. Nevertheless, as demonstrated in several works \cite{Srivastava14}, relations between different modalities may be highly complex and they cannot easily be modeled by a single layer. To account for the non-linearity in multi-modal data modeling, we recently proposed a CNN that incorporates dense connections not only between the pairs of layers within the same path, but also between those across different paths \cite{dolz2018isointense,dolz2018hyperdense}. This architecture, known as {\em HyperDenseNet}, obtained very competitive performance in the context of infant and adult brain tissue segmentation with multiple MRI data.


Despite the remarkable performance of existing methods, the combination of multi-modal data at various levels of abstraction has not been fully exploited for the segmentation of ischemic stroke lesions. In this paper, we adopt the strategy presented in \cite{dolz2018isointense,dolz2018hyperdense} and propose a multi-path architecture, where each modality is employed as input of one stream and dense connectivity is used between layers in the same and different paths. Furthermore, we also extend the standard convolutional module of InceptionNet \cite{szegedy2016rethinking} by including two additional dilated convolutional blocks, which may help to learn larger context. Experiments on 103 ischemic stroke lesion multi-modal scans from the Ischemic Stroke Lesion Segmentation (ISLES) Challenge 2018 shows our model to outperform architectures based on early and late fusion, as well as state-of-art segmentation networks.

\section{Methodology}

The proposed models build upon the UNet architecture \cite{ronneberger2015u}, which has shown outstanding performance in various medical segmentation tasks \cite{cciccek20163d,dong2017automatic}. This network consists of a contracting and expanding path, the former collapsing an image down into a set of high level features and the latter using these features to construct a pixel-wise segmentation mask. Using skip connections, outputs from early layers are concatenated to the input of subsequent layers with the objective of transferring information that may be lost in the encoding path. 

\subsection{Proposed multi-modal UNet}
\label{ssec:proposedNet}

\begin{figure}[t!]
\centering
\includegraphics[width=1\textwidth]{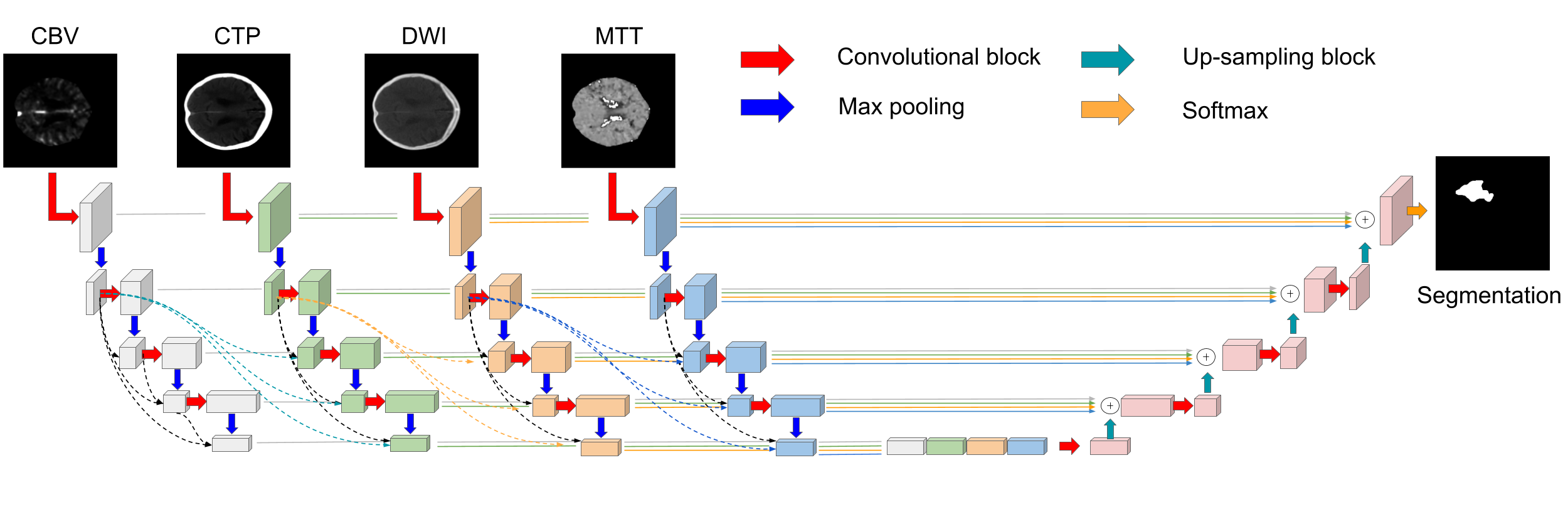}
\caption{Proposed architecture multi-path dense UNet. Dotted lines represent some of the dense connectivity patterns adopted in this extended version of UNet.} \label{HDUnet}
\label{fig:mainNet}
\end{figure}

\paragraph{\textbf{Disentangling input data.}} Figure \ref{fig:mainNet} depicts our proposed network for ischemic stroke lesion segmentation in multiple image modalities. Unlike most UNet-like architectures, the encoding path is split into $N$ streams, which serve as input to each image modality. The main objective of processing each modality in separated streams is to disentangle information that otherwise would be fused from an early stage, with the drawbacks introduced before, i.e., limitation to capture complex relationships between modalities.

\begin{figure}[h!]
\centering
\includegraphics[width=.34\textwidth]{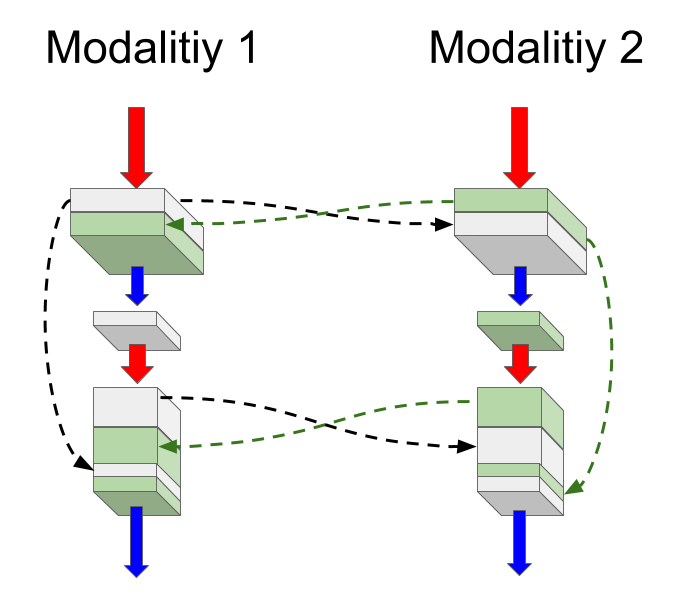}
\caption{Detailed version of a section of the proposed dense connectivity in multi-modal scenarios. For simplicity two image modalities are considered in this example.} 
\label{fig:HD_Detail}
\end{figure}

\paragraph{\textbf{Hyper-Dense connectivity.}} Inspired by the recent success of densely and hyper-densely connected networks in medical image segmentation works \cite{yu2017automatic,chen2018mri,dolz2018isointense,dolz2018hyperdense}, we propose to extend UNet to accommodate hyper-dense connections within the same and between multiple paths. In addition to better  modeling relationships between different modalities, employing dense connections also brings the three following benefits \cite{huang2017densely}. First, direct connections between all layers help improving the flow of information and gradients through the entire network, alleviating the problem of vanishing gradient. Second, short paths to all the feature maps in the architecture introduce an implicit deep supervision. And third, dense connections have a regularizing effect, which reduces the risk of over-fitting on tasks with smaller training sets.

In standard CNNs, the output of the $l^{th}$ layer, denoted as $\xx_l$, is typically obtained from the output of the previous layer $\xx_{l-1}$ by a mapping $H_l$:
\begin{equation}
 \xx_l \ = \ H_l\big(\xx_{l-1}\big).
 \label{eq:layer_output}
\end{equation} 
where $H_l$ commonly integrates a convolution layers followed by a non-linear activation. In a densely-connected network, all feature outputs are concatenated in a feed-forward manner,
\begin{equation}
 \xx_l \ = \ H_l\big([\xx_{l-1}, \xx_{l-2}, \ldots, \xx_{0}]\big),
 \label{eq:layer_outputDense}
\end{equation}
where $[\ldots]$ denotes a concatenation operation. 

As in HyperDenseNet \cite{dolz2018isointense,dolz2018hyperdense}, the outputs from layers in different streams are also linked. This connectivity yields a much more powerful feature representation than early or late fusion strategies in a multi-modal context, as the network learns the complex relationships between the modalities within and in-between all the levels of abstractions. Considering the case of only two modalities, let $\xx_l^1$ and $\xx_l^2$ denote the outputs of the $l^{th}$ layer in streams 1 and 2, respectively. In general, the output of the $l^{th}$ layer in a stream $s$ can then be defined as follows:
\begin{equation}
 \xx_l^s \ = \ H_l^s\big([\xx_{l-1}^1, \xx_{l-1}^2, \xx_{l-2}^1, \xx_{l-2}^2, \ldots, \xx_{0}^1, \xx_{0}^2]\big).
 \label{eq:layer_HyperDense}
\end{equation}

Inspired by the recent findings in \cite{chen2017regularization,zhang2017interleaved,zhang2017shufflenet}, where shuffling and interleaving feature map elements in a CNN improved the efficiency and performance, while serving as a strong regularizer, we concatenate feature maps in a different order for each branch and layer:
\begin{equation}
  \xx_l^s \ = \ H_l^s\big(\pi_l^s([\xx_{l-1}^1, \xx_{l-1}^2, \xx_{l-2}^1, \xx_{l-2}^2, \ldots, \xx_{0}^1, \xx_{0}^2])\big),
\end{equation}
with $\pi_l^s$ being a function that permutes the feature maps given as input. Thus, in the case of two image modalities, we have: 
\begin{equation*}
\begin{split}
  \xx_l^1 & \ = \ 
    H_l^1\big([\xx_{l-1}^1, \xx_{l-1}^2, \xx_{l-2}^1, \xx_{l-2}^2, \ldots, \xx_{0}^1, \xx_{0}^2]\big) \\ 
  \xx_l^2 & \ = \ 
    H_l^2\big([\xx_{l-1}^2, \xx_{l-1}^1, \xx_{l-2}^2, \xx_{l-2}^1, \ldots, \xx_{0}^2, \xx_{0}^1])\big.
  \end{split}
\end{equation*}

A detailed example of hyper-dense connectivity for the case of two image modalities is depicted in Fig. \ref{fig:HD_Detail}.

\subsection{Extended Inception module}

Salient regions in a given image can have extremely large variation in size. For example, in ischemic stroke lesion segmentation, the area occupied by a lesion highly varies from one image to another. Therefore, choosing the appropriate kernel size is not trivial. While a smaller kernel is better for local information, a larger kernel is preferred to capture information that is distributed globally. InceptionNet \cite{szegedy2016rethinking} exploits this principle by including convolutions with multiple kernel sizes which operate on the same level. Furthermore, in versions 2 and 3, convolutions of the shape n$\times$n are factorized to a combination of 1$\times$n and n$\times$1 convolutions, which have demonstrated to be more efficient. For example, a 3$\times$3 convolution is equivalent to a 1$\times$3 followed by a 3$\times$1 convolution, which was found to be 33$\%$ cheaper.

\begin{figure}[h!]
\centering
\mbox{
\includegraphics[width=0.5\textwidth,trim={2.3cm 0 2.5cm 0},clip]{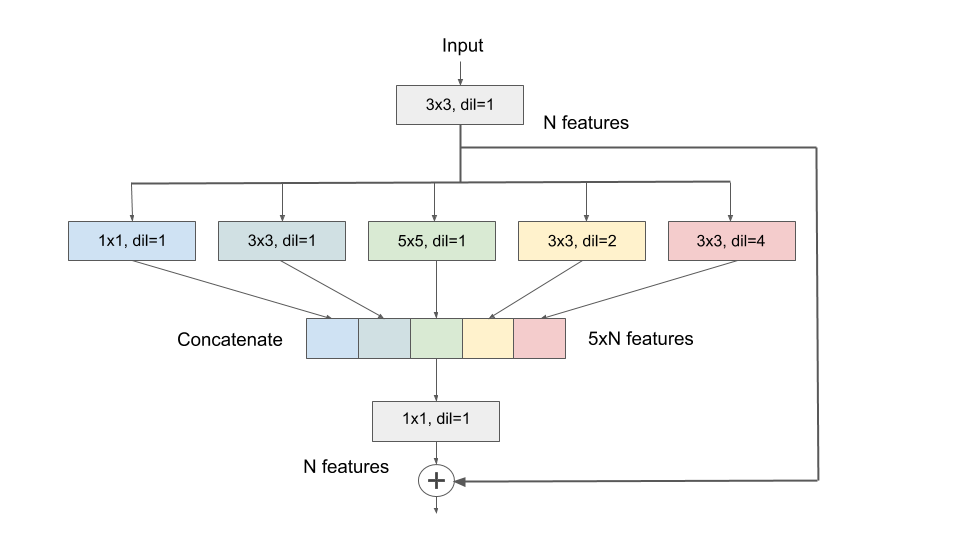}
\includegraphics[width=0.5\textwidth,trim={2.3cm 0 2.5cm 0},clip]{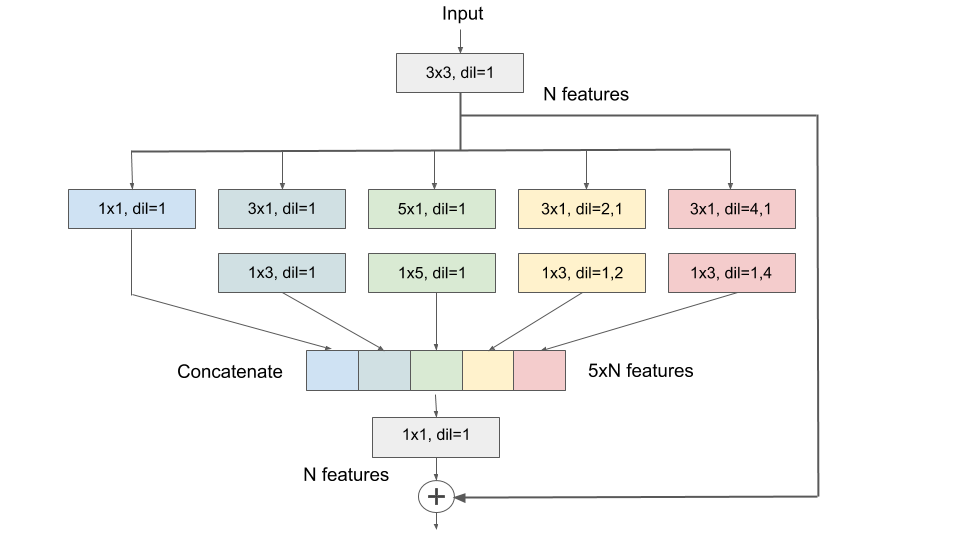}
}
\caption{Proposed extended inception modules. The module on the left employs standard convolutions while the module on the right adopts the idea of asymmetric convolutions \cite{szegedy2016rethinking}.}
\label{fig:res-inception-mod}
\end{figure}

We also extended the convolutional module of InceptionNet to facilitate the learning of multiple context. Particularly, we included two additional convolutional blocks, with different dilation rates, which help the module to learn from multiple receptive fields and to increase the context with respect to the original inception module. Since dilated convolutions were shown to be better alternative to max-pooling when capturing global context \cite{yu2015multi}, we removed the latter operation in the proposed module. Our extended inception modules are depicted in Fig. \ref{fig:res-inception-mod}.

\section{Materials}

\subsection{Dataset}

The training dataset, composed of 103 ischemic stroke lesion multi-modal scans, was provided by the ISLES organizers. We split the 94 stroke datasets into training and validation sets containing 83 and 11 examples, respectively. Each scan contains: Diffusion maps (DWI) and Perfusion maps (CBF,MTT,CBV,Tmax and CTP source data). In addition, the manual ground truth segmentation is provided only for the training samples. Detailed information about the dataset can be found in the ISLES website\footnote{http://www.isles-challenge.org}.

\subsection{Evaluation metrics}

\paragraph{\textbf{Dice similarity coefficient (DSC).}} We first evaluate performance using Dice similarity coefficient (DSC), which compares volumes based on their overlap. Let $V_\mr{ref}$ and $V_\mr{auto}$ be, respectively, the reference and automatic segmentations of a given tissue class and for a given subject, the DSC for this subject is defined as:
\begin{equation}
\label{eq:dice}
\mr{DSC}\big(V_\mr{ref}, V_\mr{auto} \big) \ = \ 
 \frac{2 \mid V_\mr{ref} \cap V_\mr{auto}\mid} {\mid V_\mr{ref}\mid +\mid V_\mr{auto}\mid}
\end{equation}

\paragraph{\textbf{Modified Hausdorff distance (MHD).}} The second metric measures the accuracy of the segmentation boundary. Let $P_\mr{ref}$ and $P_\mr{auto}$ denote the sets of voxels within the reference and automatic segmentation boundary, respectively. MHD is given by
\begin{small}
\begin{equation}
\label{eq:hausd}
\mr{MHD}\big(P_\mr{ref}, P_\mr{auto} \big) \ = \ \max \Big\{ \max_{q \in P_\mr{ref}}d(q,P_\mr{auto}), \max_{q \in P_\mr{auto}}d(q,P_\mr{ref}) \Big\},
\end{equation}
\end{small}
where $d(q,P)$ is the point-to-set distance defined by: $d(q,P)=\min_{p \in P} \| q-p\|$, with $\|.\|$ denoting the Euclidean distance. In the MHD, the 95$^{th}$ percentile is used for the estimation of the maximum distance value. Low MHD values indicate high boundary similarity. 

\paragraph{\textbf{Volumetric similarity (VS).}} Volumetric similarity (VS) ignores the overlap between the predicted and reference segmentations, and simply compares the size of the predicted volume to that of the reference:
\begin{equation} \mathrm{VS(V_\mr{ref},V_\mr{auto})} \ = \ 1- \dfrac{\left|\ |V_\mr{ref}| \,-\, |V_\mr{auto}|\ \right|}{|V_\mr{ref}| \,+\, |V_\mr{auto}|}. 
\end{equation}
where a VS equal to 1 reflects that the predicted segmentation is the same size as the reference volume.

\subsection{Implementation details}

\paragraph{\textbf{\textit{Baselines.}}} To demonstrate the effectiveness of hyper-dense connectivity in deep neural networks we compare the proposed architecture to the same network with early and late fusion strategies. For early fusion, all the MRI image modalities are merged into a single input, which is processed through a unique path, as many current works. On the other hand, each image modality is treated as an independent signal and processed by separate branches in the later fusion strategy, where features are fused at a higher level. The details of the late fusion architecture are depicted in Table \ref{ref:arch}. In both cases, i.e., early and late fusion, the left module depicted in Fig. \ref{fig:res-inception-mod} is employed. Furthermore, feature maps from the skip connections are summed before being fed into the convolutional modules of the decoding path, instead of concatenating them, as in standard UNet.

\paragraph{\textbf{\textit{Proposed network.}}} The proposed network is similar to the architecture with the late fusion strategy. Nevertheless, as introduced in section \ref{ssec:proposedNet}, feature maps from previous layers and different paths are concatenated and fed into the subsequent layers. The details of the resulted architecture are reported in Table \ref{ref:arch}, most-right columns. The first version of the proposed network employs the same convolutional module than the two baselines. The second version, however, adopts asymmetric convolutions instead (Fig. \ref{fig:res-inception-mod}).

\begin{table}[]
\centering
\scriptsize
\caption{Layers disposal of the architecture with late fusion and the proposed hyper dense connected UNet.}
\begin{tabular}{|l|l||c|c||c|c|}
\hline
\multicolumn{2}{|l||}{} & \multicolumn{2}{c||}{\textbf{Late fusion}} & \multicolumn{2}{c|}{\textbf{HyperDense connectivity}} \\  
\hline &   Name & \multicolumn{1}{c|}{\begin{tabular}[c]{@{}c@{}}Feat maps\\ (input)\end{tabular}} & \multicolumn{1}{c||}{\begin{tabular}[c]{@{}c@{}}Feat maps\\ (output)\end{tabular}} & \multicolumn{1}{c|}{\begin{tabular}[c]{@{}c@{}}Feat maps\\ (input)\end{tabular}} & \multicolumn{1}{c|}{\begin{tabular}[c]{@{}c@{}}Feat maps\\ (output)\end{tabular}} \\ \hline
\multirow{7}{*}{\begin{tabular}[c]{@{}l@{}}Encoding\\ Path \\ (each modality)\end{tabular}} 
& Conv Layer 1       & 1$\times$256$\times$256 &32$\times$256$\times$256 & 1$\times$256$\times$256 &32$\times$256$\times$256 \\ 
& Max-pooling 1       & 32$\times$256$\times$256 & 32$\times$128$\times$128 & 32$\times$256$\times$256 & 32$\times$128$\times$128 \\
& Layer 2       &  32$\times$128$\times$128 & 64$\times$128$\times$128 &  128$\times$128$\times$128 & 64$\times$128$\times$128  \\
& Max-pooling 2       & 64$\times$128$\times$128  & 64$\times$64$\times$64  & 64$\times$128$\times$128  & 64$\times$64$\times$64  \\
& Layer 3       & 64$\times$64$\times$64 & 128$\times$64$\times$64 & 384$\times$64$\times$64 & 128$\times$64$\times$64   \\
& Max-pooling 3       & 128$\times$64$\times$64 & 128$\times$32$\times$32  & 128$\times$64$\times$64 & 128$\times$32$\times$32 \\
& Layer 4       & 128$\times$32$\times$32 &  256$\times$32$\times$32  & 896$\times$32$\times$32 &  256$\times$32$\times$32  \\ 
& Max-pooling 4       & 256$\times$32$\times$32 &  256$\times$16$\times$16 & 256$\times$32$\times$32 &  256$\times$16$\times$16 \\\hline
 &   Bridge  & 1024$\times$16$\times$16 &  512$\times$16$\times$16 & 1920$\times$16$\times$16 &  512$\times$16$\times$16 \\\hline
\multirow{5}{*}{\begin{tabular}[c]{@{}l@{}}Decoding\\ Path\end{tabular}} 
& Up-sample 1       & 512$\times$16$\times$16 & 256$\times$32$\times$32  & 512$\times$16$\times$16 & 256$\times$32$\times$32 \\
& Layer 5       & 256$\times$32$\times$32  &  256$\times$32$\times$32 & 256$\times$32$\times$32  &  256$\times$32$\times$32  \\
& Up-sample 2    & 256$\times$32$\times$32 & 128$\times$64$\times$64  & 256$\times$32$\times$32 & 128$\times$64$\times$64   \\
& Layer 6    & 128$\times$64$\times$64  &  128$\times$64$\times$64 & 128$\times$64$\times$64  &  128$\times$64$\times$64  \\
& Up-sample 3       & 128$\times$64$\times$64 &  64$\times$128$\times$128 & 128$\times$64$\times$64 &  64$\times$128$\times$128 \\
& Layer 7       & 64$\times$128$\times$128 &  64$\times$128$\times$128 & 64$\times$128$\times$128 &  64$\times$128$\times$128 \\
& Up-sample 4       & 64$\times$128$\times$128 &  32$\times$256$\times$256 & 64$\times$128$\times$128 &  32$\times$256$\times$256 \\
& Layer 8       & 32$\times$256$\times$256 & 32$\times$256$\times$256 & 32$\times$256$\times$256 & 32$\times$256$\times$256   \\
& Softmax layer &32$\times$256$\times$256  & 2$\times$256$\times$256 &32$\times$256$\times$256  & 2$\times$256$\times$256  \\ \hline
\end{tabular}
\label{ref:arch}
\end{table}

\paragraph{\textbf{\textit{Training.}}} Network parameters were optimized via Adam with $\beta_1$ and $\beta_2$ equal to 0.9 and 0.99, respectively and training is run during 200 epochs. Learning rate is initially set to 1$\times$10$^{-4}$ and reduced after 100 epochs. Batch size was equal to 4. For a fair comparison, the same hyper-parameters were employed across all the architectures. The proposed architectures were implemented in pytorch. Experiments were performed on a NVidia TITAN XP GPU with 16 GBs RAM. While training took around 5 hours, inference on a single 2D image was done in 0.1 sec, as average. No data augmentation was employed. Images were normalized between 0 and 1 and no other pre- or post-processing steps were used. As input to the architectures we employed the following four image modalities in all the cases: CBV, CTP, DWI and MTT.

\vspace{-0.2cm}

\section{Results}

Table \ref{table:metrics} reports the results obtained by the different networks that we investigated in terms of mean DSC, MHD and VS values and their standard deviation. First, we compare the different multi-modal fusion strategies with the baseline UNet employed in this work. We can observe that fusing learning features in a higher level provides better results in all the metrics than early fusion strategies. Additionally, if hyper-dense connections are adopted in the late fusion architecture, i.e., interconnecting convolutional layers from the different image modalities, the segmentation performance is significantly improved, particularly in terms of DSC and VS. Specifically, while the proposed network outperforms the late fusion architecture by nearly 5$\%$ in both DSC and VS, the mean MHD is decreased by almost 1 mm, obtaining a mean MHD of 18.88 mm. On the other hand, replacing the standard convolutions of the proposed module (Fig \ref{fig:res-inception-mod},\textit{left}) by asymmetric convolutions (Fig \ref{fig:res-inception-mod},\textit{right}), brings another boost on performance on the proposed hyper-dense UNet. In this case, the mean DSC and MHD are the best ones among all the architectures, with mean values of 0.635$\%$ and 18.64 mm, respectively. 

\begin{table}[]

\centering
\scriptsize
\caption{Mean DSC,MHD and VS values, with their corresponding standard deviation, obtained by the evaluated methods on the independent validation group.}
\begin{tabular}{lccc}
\toprule
 & \multicolumn{3}{c}{\textbf{Validation}}           \\ \toprule
Architecture & DSC ($\%$)        & MHD (mm)       & VS ($\%$)         \\ \midrule
Early Fusion   & 0.497$\,\pm\,$0.263  &
21.30$\,\pm\,$13.25 & 0.654$\,\pm\,$0.265  \\
Late Fusion &0.571$\,\pm\,$0.221  &
19.72$\,\pm\,$12.29 & 0.718$\,\pm\,$0.235\\
Proposed              & 0.622$\,\pm\,$0.233  &
18.88$\,\pm\,$14.87 & 0.764$\,\pm\,$0.247 \\  
Proposed (Asymmetric conv)              & \textbf{0.635}$\,\pm\,$0.186  &
\textbf{18.64}$\,\pm\,$14.26 & 0.796$\,\pm\,$0.162\\  \midrule
ERFNet \cite{romera2018erfnet}            &  0.540$\,\pm\,$0.258&\hspace{1em} 21.73$\,\pm\,$11.46 \hspace{1em}&  \textbf{0.823}$\,\pm\,$0.119 \\
\bottomrule

\end{tabular}
\label{table:metrics}

\end{table}

Then, we also compare the results obtained by the proposed network to another state-of-the-art network that includes factorized convolution modules, i.e., ERFNet. Even though its performance outperforms the baseline with early fusion, results are far from those obtained by the proposed network, except for the volume similarity, where both ERFNet and the proposed network with asymmetric convolutions obtain similar performances. 



\begin{figure}[h!]
\centering
\includegraphics[width=1\textwidth]{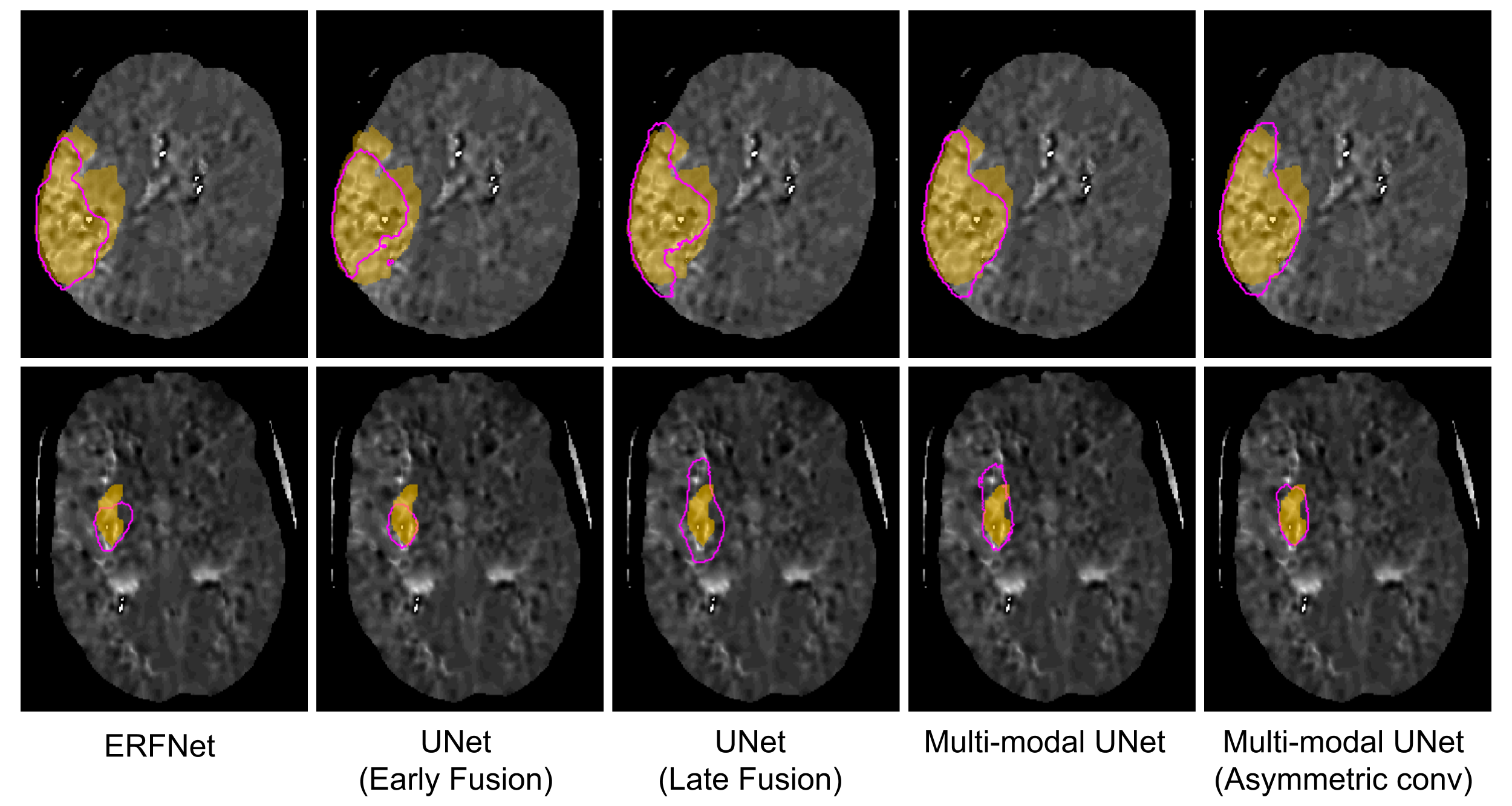}
\caption{Visual results for two subjects on the validation set. While the area in yellow represents the ground truth, purple contours depict the automatic contours for each of the architectures.} 
\label{fig:ISLES_Res}
\end{figure}

Qualitative evaluation of the proposed architecture is assessed in Fig. \ref{fig:ISLES_Res}, where ground truth and automatic CNN contours are visualized on MTT images. We can first observe that, by employing strategies where learned features are merged at higher levels, unlike \textit{early fusion}, the region of the ischemic stroke lesion is generally better covered. Furthermore, by giving freedom to the architecture to learn the level of abstraction at which the different modalities should be combined segmentation results are visually improved, which is in line with the results reported in Table \ref{table:metrics}.


\vspace{-1cm}
\section{Discussion}

In this work we extended the well-known UNet to leverage information in multi-modal data. Particularly, following recent work in multi-modal learning for medical image segmentation \cite{dolz2018isointense,dolz2018hyperdense}, we processed each image modality in different streams on the encoding path and densely connected all the convolutional layers from all the streams. Thus, each convolutional layer received as input the features maps from all the previous layers within the same stream, i.e., same modality, but also the learned features from previous layers in every different stream. In this way, the network has the freedom to learn any pattern at any level of abstraction of the network, which seems to improve its representation learning power.  

Results obtained in this work demonstrate that better strategies to model multi-modal information can bring a boost on performance when compared to more naive fusion strategies. These results are in line with recent studies in multi-modal image segmentation on the medical field \cite{nie2016fully,dolz2018isointense,dolz2018hyperdense}. For instance, in \cite{nie2016fully}, a {\em late fusion} strategy was proposed to combine high-level features to better capture the complex relationships between different modalities. They used an independent convolutional network for each modality, and fused the outputs of the different networks in higher-level layers, showing better performance than early fusion in the context infant brain segmentation. More recently, we demonstrated that hyper dense connectivity can strength the representation power of deep CNNs in the context of multi-modal image infant and adult brain segmentation, surpassing the performance of several features fusion strategies \cite{dolz2018isointense,dolz2018hyperdense}.

One of the limitations of this work is that volumes were treated as a stack of 2D slides, where each slide was processed independently. Thus, 3D context was discarded, which might have improved the segmentation performance, as shown by recent works that employ 3D convolutions. One of the reasons for privileging 2D convolutions is that some of the volumes on the ISLES dataset contained a limited number of slides, i.e., 2 and 4 slides in many cases. One strategy to explore in the future could be to employ Long-Short Term Memory (LSTM) networks to propagate the spatial information extracted from the 2D CNN through the third dimension. 

\subsubsection*{Acknowledgments}

This work is supported by the National Science and Engineering Research Council of Canada (NSERC), discovery grant program, and by the ETS Research Chair on Artificial Intelligence in Medical Imaging. 


\bibliographystyle{abbrv}
{
\begin{footnotesize}
\bibliography{biblio}

\begin{thebibliography}{10}

\bibitem{aygun2018multi}
M.~Ayg{\"u}n, Y.~H. {\c{S}}ahin, and G.~{\"U}nal.
\newblock Multi modal convolutional neural networks forbrain tumor
  segmentation.
\newblock {\em arXiv preprint arXiv:1809.06191}, 2018.

\bibitem{barber2005imaging}
P.~Barber, M.~Hill, M.~Eliasziw, A.~Demchuk, J.~Pexman, M.~Hudon, A.~Tomanek,
  R.~Frayne, and A.~Buchan.
\newblock Imaging of the brain in acute ischaemic stroke: comparison of
  computed tomography and magnetic resonance diffusion-weighted imaging.
\newblock {\em Journal of Neurology, Neurosurgery \& Psychiatry},
  76(11):1528--1533, 2005.

\bibitem{chalela2007magnetic}
J.~A. Chalela, C.~S. Kidwell, L.~M. Nentwich, M.~Luby, J.~A. Butman, A.~M.
  Demchuk, M.~D. Hill, N.~Patronas, L.~Latour, and S.~Warach.
\newblock Magnetic resonance imaging and computed tomography in emergency
  assessment of patients with suspected acute stroke: a prospective comparison.
\newblock {\em The Lancet}, 369(9558):293--298, 2007.

\bibitem{chen2017fully}
L.~Chen, P.~Bentley, and D.~Rueckert.
\newblock Fully automatic acute ischemic lesion segmentation in {DWI} using
  convolutional neural networks.
\newblock {\em NeuroImage: Clinical}, 15:633--643, 2017.

\bibitem{chen2018mri}
L.~Chen, Y.~Wu, A.~M. DSouza, A.~Z. Abidin, A.~Wism{\"u}ller, and C.~Xu.
\newblock Mri tumor segmentation with densely connected {3D} {CNN}.
\newblock In {\em Medical Imaging 2018: Image Processing}. International
  Society for Optics and Photonics, 2018.

\bibitem{chen2017regularization}
Y.~Chen, H.~Wang, and Y.~Long.
\newblock Regularization of convolutional neural networks using shufflenode.
\newblock In {\em Multimedia and Expo (ICME), 2017 IEEE International
  Conference on}, pages 355--360. IEEE, 2017.

\bibitem{choi2016ensemble}
Y.~Choi, Y.~Kwon, H.~Lee, B.~J. Kim, M.~C. Paik, and J.-H. Won.
\newblock Ensemble of deep convolutional neural networks for prognosis of
  ischemic stroke.
\newblock In {\em Workshop on Brainlesion: Glioma, Multiple Sclerosis, Stroke
  and Traumatic Brain Injuries}, pages 231--243. Springer, 2016.

\bibitem{cciccek20163d}
{\"O}.~{\c{C}}i{\c{c}}ek, A.~Abdulkadir, S.~S. Lienkamp, T.~Brox, and
  O.~Ronneberger.
\newblock 3d u-net: learning dense volumetric segmentation from sparse
  annotation.
\newblock In {\em MICCAI}, pages 424--432. Springer, 2016.

\bibitem{dolz2018isointense}
J.~Dolz, I.~Ben~Ayed, J.~Yuan, and C.~Desrosiers.
\newblock Isointense infant brain segmentation with a hyper-dense connected
  convolutional neural network.
\newblock In {\em Biomedical Imaging (ISBI 2018), 2018 IEEE 15th International
  Symposium on}, pages 616--620. IEEE, 2018.

\bibitem{dolz2017deep}
J.~Dolz, C.~Desrosiers, L.~Wang, J.~Yuan, D.~Shen, and I.~B. Ayed.
\newblock Deep {CNN} ensembles and suggestive annotations for infant brain
  {MRI} segmentation.
\newblock {\em arXiv preprint arXiv:1712.05319}, 2017.

\bibitem{dolz2018hyperdense}
J.~Dolz, K.~Gopinath, J.~Yuan, H.~Lombaert, C.~Desrosiers, and I.~B. Ayed.
\newblock Hyperdense-net: A hyper-densely connected cnn for multi-modal image
  segmentation.
\newblock {\em arXiv preprint arXiv:1804.02967}, 2018.

\bibitem{dong2017automatic}
H.~Dong, G.~Yang, F.~Liu, Y.~Mo, and Y.~Guo.
\newblock Automatic brain tumor detection and segmentation using u-net based
  fully convolutional networks.
\newblock In {\em MIUA}, pages 506--517. Springer, 2017.

\bibitem{feigin2003stroke}
V.~L. Feigin, C.~M. Lawes, D.~A. Bennett, and C.~S. Anderson.
\newblock Stroke epidemiology: a review of population-based studies of
  incidence, prevalence, and case-fatality in the late 20th century.
\newblock {\em The Lancet Neurology}, 2(1):43--53, 2003.

\bibitem{guerrero2018white}
R.~Guerrero, C.~Qin, O.~Oktay, C.~Bowles, L.~Chen, R.~Joules, R.~Wolz,
  M.~Vald{\'e}s-Hern{\'a}ndez, D.~Dickie, J.~Wardlaw, et~al.
\newblock White matter hyperintensity and stroke lesion segmentation and
  differentiation using convolutional neural networks.
\newblock {\em NeuroImage: Clinical}, 17:918--934, 2018.

\bibitem{huang2017densely}
G.~Huang, Z.~Liu, L.~Van Der~Maaten, and K.~Q. Weinberger.
\newblock Densely connected convolutional networks.
\newblock In {\em CVPR}, volume~1, page~3, 2017.

\bibitem{kamnitsas2015multi}
K.~Kamnitsas, L.~Chen, C.~Ledig, D.~Rueckert, and B.~Glocker.
\newblock Multi-scale {3D} convolutional neural networks for lesion
  segmentation in brain {MRI}.
\newblock {\em Ischemic Stroke Lesion Segmentation}, 13, 2015.

\bibitem{kamnitsas2017efficient}
K.~Kamnitsas, C.~Ledig, V.~F. Newcombe, J.~P. Simpson, A.~D. Kane, D.~K. Menon,
  D.~Rueckert, and B.~Glocker.
\newblock Efficient multi-scale {3D} {CNN} with fully connected {CRF} for
  accurate brain lesion segmentation.
\newblock {\em Medical image analysis}, 36:61--78, 2017.

\bibitem{lansberg2000comparison}
M.~G. Lansberg, G.~W. Albers, C.~Beaulieu, and M.~P. Marks.
\newblock Comparison of diffusion-weighted mri and ct in acute stroke.
\newblock {\em Neurology}, 54(8):1557--1561, 2000.

\bibitem{lopez2006global}
A.~D. Lopez, C.~D. Mathers, M.~Ezzati, D.~T. Jamison, and C.~J. Murray.
\newblock Global and regional burden of disease and risk factors, 2001:
  systematic analysis of population health data.
\newblock {\em The Lancet}, 367(9524):1747--1757, 2006.

\bibitem{maier2017isles}
O.~Maier, B.~H. Menze, J.~von~der Gablentz, L.~H{\"a}ni, M.~P. Heinrich,
  M.~Liebrand, S.~Winzeck, A.~Basit, P.~Bentley, L.~Chen, et~al.
\newblock Isles 2015-a public evaluation benchmark for ischemic stroke lesion
  segmentation from multispectral {MRI}.
\newblock {\em Medical image analysis}, 35:250--269, 2017.

\bibitem{maier2015classifiers}
O.~Maier, C.~Schr{\"o}der, N.~D. Forkert, T.~Martinetz, and H.~Handels.
\newblock Classifiers for ischemic stroke lesion segmentation: a comparison
  study.
\newblock {\em PloS one}, 10(12):e0145118, 2015.

\bibitem{moeskops2016automatic}
P.~Moeskops, M.~A. Viergever, A.~M. Mendrik, L.~S. de~Vries, M.~J. Benders, and
  I.~I{\v{s}}gum.
\newblock Automatic segmentation of {MR} brain images with a convolutional
  neural network.
\newblock {\em IEEE Transactions on Medical Imaging}, 35(5):1252--1261, 2016.

\bibitem{nie2016fully}
D.~Nie, L.~Wang, Y.~Gao, and D.~Sken.
\newblock Fully convolutional networks for multi-modality isointense infant
  brain image segmentation.
\newblock In {\em 13th International Symposium on Biomedical Imaging (ISBI),
  2016}, pages 1342--1345. IEEE, 2016.

\bibitem{praveen2018ischemic}
G.~Praveen, A.~Agrawal, P.~Sundaram, and S.~Sardesai.
\newblock Ischemic stroke lesion segmentation using stacked sparse autoencoder.
\newblock {\em Computers in biology and medicine}, 2018.

\bibitem{rekik2012medical}
I.~Rekik, S.~Allassonni{\`e}re, T.~K. Carpenter, and J.~M. Wardlaw.
\newblock Medical image analysis methods in {MR/CT}-imaged acute-subacute
  ischemic stroke lesion: {S}egmentation, prediction and insights into dynamic
  evolution simulation models. {A} critical appraisal.
\newblock {\em NeuroImage: Clinical}, 1(1):164--178, 2012.

\bibitem{romera2018erfnet}
E.~Romera, J.~M. Alvarez, L.~M. Bergasa, and R.~Arroyo.
\newblock Erfnet: Efficient residual factorized convnet for real-time semantic
  segmentation.
\newblock {\em IEEE Transactions on Intelligent Transportation Systems},
  19(1):263--272, 2018.

\bibitem{ronneberger2015u}
O.~Ronneberger, P.~Fischer, and T.~Brox.
\newblock U-net: Convolutional networks for biomedical image segmentation.
\newblock In {\em MICCAI}, pages 234--241. Springer, 2015.

\bibitem{seshadri2007lifetime}
S.~Seshadri and P.~A. Wolf.
\newblock Lifetime risk of stroke and dementia: current concepts, and estimates
  from the framingham study.
\newblock {\em The Lancet Neurology}, 6(12):1106--1114, 2007.

\bibitem{sirinukunwattana2017gland}
K.~Sirinukunwattana, J.~P. Pluim, H.~Chen, X.~Qi, P.-A. Heng, Y.~B. Guo, L.~Y.
  Wang, B.~J. Matuszewski, E.~Bruni, U.~Sanchez, et~al.
\newblock Gland segmentation in colon histology images: The glas challenge
  contest.
\newblock {\em Medical image analysis}, 35:489--502, 2017.

\bibitem{Srivastava14}
N.~Srivastava and R.~Salakhutdinov.
\newblock Multimodal learning with deep boltzmann machines.
\newblock {\em Journal of Machine Learning Research}, 15:2949--2980, 2014.

\bibitem{sudlow1997comparable}
C.~Sudlow and C.~Warlow.
\newblock Comparable studies of the incidence of stroke and its pathological
  types: results from an international collaboration.
\newblock {\em Stroke}, 28(3):491--499, 1997.

\bibitem{szegedy2016rethinking}
C.~Szegedy, V.~Vanhoucke, S.~Ioffe, J.~Shlens, and Z.~Wojna.
\newblock Rethinking the inception architecture for computer vision.
\newblock In {\em CVPR}, pages 2818--2826, 2016.

\bibitem{valverde2017improving}
S.~Valverde, M.~Cabezas, E.~Roura, S.~Gonz{\'a}lez-Vill{\`a}, D.~Pareto, J.~C.
  Vilanova, L.~Rami{\'o}-Torrent{\`a}, {\`A}.~Rovira, A.~Oliver, and
  X.~Llad{\'o}.
\newblock Improving automated multiple sclerosis lesion segmentation with a
  cascaded {3D} convolutional neural network approach.
\newblock {\em NeuroImage}, 155:159--168, 2017.

\bibitem{van2007acute}
H.~B. Van~der Worp and J.~van Gijn.
\newblock Acute ischemic stroke.
\newblock {\em New England Journal of Medicine}, 357(6):572--579, 2007.

\bibitem{winzeck2018isles}
S.~Winzeck, A.~Hakim, R.~McKinley, J.~A. Pinto, V.~Alves, C.~Silva, M.~Pisov,
  E.~Krivov, M.~Belyaev, M.~Monteiro, et~al.
\newblock Isles 2016 and 2017-benchmarking ischemic stroke lesion outcome
  prediction based on multispectral mri.
\newblock {\em Frontiers in Neurology}, 9, 2018.

\bibitem{yu2015multi}
F.~Yu and V.~Koltun.
\newblock Multi-scale context aggregation by dilated convolutions.
\newblock {\em arXiv preprint arXiv:1511.07122}, 2015.

\bibitem{yu2017automatic}
L.~Yu, J.-Z. Cheng, Q.~Dou, X.~Yang, H.~Chen, J.~Qin, and P.-A. Heng.
\newblock Automatic {3D} cardiovascular {MR} segmentation with
  densely-connected volumetric convnets.
\newblock In {\em MICCAI}, pages 287--295. Springer, 2017.

\bibitem{zhang2017interleaved}
T.~Zhang, G.-J. Qi, B.~Xiao, and J.~Wang.
\newblock Interleaved group convolutions.
\newblock In {\em CVPR}, pages 4373--4382, 2017.

\bibitem{zhang2015deep}
W.~Zhang, R.~Li, H.~Deng, L.~Wang, W.~Lin, S.~Ji, and D.~Shen.
\newblock Deep convolutional neural networks for multi-modality isointense
  infant brain image segmentation.
\newblock {\em NeuroImage}, 108:214--224, 2015.

\bibitem{zhang2017shufflenet}
X.~Zhang, X.~Zhou, M.~Lin, and J.~Sun.
\newblock Shufflenet: An extremely efficient convolutional neural network for
  mobile devices.
\newblock {\em arXiv preprint arXiv:1707.01083}, 2017.

\end{thebibliography}
\end{footnotesize}
}

\end{document}